\pdfoutput=1

\documentclass[11pt]{article}

\usepackage{ACL2023}

\usepackage{times}
\usepackage{latexsym}
\usepackage{amsmath}
\usepackage{graphicx}
\usepackage{tabularx}
\usepackage{listings}
\usepackage[T1]{fontenc}

\usepackage[utf8]{inputenc}

\usepackage{microtype}

\usepackage{CJKutf8}

\usepackage{subcaption} 

\usepackage[symbol]{footmisc}

\newcommand{\loss}{\ell}
\newcommand{\eval}{$D_{eval}$}
\newcommand{\full}{$D_{full}$}
\title{In2Core: 
Leveraging Influence Functions for Coreset Selection in Instruction Finetuning of Large Language Models}

\author{
\textbf{
Ayrton San Joaquin\textsuperscript{$\heartsuit$} 
\ \quad
Bin Wang\textsuperscript{$\diamondsuit$} 
\ \quad 
Zhengyuan Liu\textsuperscript{$\diamondsuit$} 
\ \quad 
}
\\
\textbf{
Nicholas Asher\textsuperscript{$\S,\heartsuit$} 
\ \quad 
Brian Lim\textsuperscript{$\infty,\heartsuit$} 
\ \quad 
Philippe Muller\textsuperscript{$\S,\heartsuit$} 
\ \quad 
Nancy F. Chen\textsuperscript{$\diamondsuit,\heartsuit,\dag$}} \\
\textsuperscript{$\heartsuit$}CNRS@CREATE, Singapore
\quad
\textsuperscript{$\diamondsuit$}Institute for Infocomm Research (I$^2$R), A*STAR, Singapore\\
\textsuperscript{$\S$}CNRS, IRIT, France
\quad
\textsuperscript{$\infty$}National University of Singapore, Singapore\\
\textsuperscript{$\dag$}Centre for Frontier AI Research (CFAR), A*STAR, Singapore
}

\begin{document}
\maketitle
\begin{abstract}
Despite advancements, fine-tuning Large Language Models (LLMs) remains costly due to the extensive parameter count and substantial data requirements for model generalization. Accessibility to computing resources remains a barrier for the open-source community. To address this challenge, we propose the \texttt{In2Core} algorithm, which selects a coreset by analyzing the correlation between training and evaluation samples with a trained model. Notably, we assess the model's internal gradients to estimate this relationship, aiming to rank the contribution of each training point. To enhance efficiency, we propose an optimization to compute influence functions with a reduced number of layers while achieving similar accuracy. By applying our algorithm to instruction fine-tuning data of LLMs, we can achieve similar performance with just 50\% of the training data. Meantime, using influence functions to analyze model coverage to certain testing samples could provide a reliable and interpretable signal on the training set's coverage of those test points.
\end{abstract}

\section{Introduction}

    With the advent of Large Language Models (LLMs) exhibiting surprising abilities across a variety of language tasks \cite{clark2018think,zellers2019hellaswag,DBLP:journals/corr/abs-1907-10641,hendrycks2021measuring,DBLP:journals/corr/abs-2110-14168,eval-harness,lin2022truthfulqa,open-llm-leaderboard}, open-source language models have surged in popularity as part of broader efforts to democratize their accessibility~\cite{alpaca}. Furthermore, performance can be improved with custom data that is smaller than their pretraining data, which is known as supervised fine-tuning~\cite{liu2024understanding}. One type of fine-tuning important to the open-source community is instruction tuning, which allows models to follow a broad or specific-set of instructions and to discuss with users in natural dialogue~\cite{zhang2023instruction}. However, instruction-tuning of open-source LLMs remains limited in multiple areas. 

    One prominent limitation is the expensive cost of fine-tuning such models, caused by a large number of parameters and data required ~\cite{liu2024datasets}. The popular approach is to scale-up the parameter count and the dataset size, increasing the computation required to train the models. This method to improve LLMs does not lend-well to organizations without massive computing resources, such as the majority of entities who rely on open-source. Another limitation is evaluation, especially for instruction-following models. Evaluation is challenging as a result of LLMs' inherent open-ended generation, where a space of ideal outputs exists, and existing evaluations usually measure ability in a specific area, such as summarization. Furthermore, evaluation sets may be hard to construct for domain-specific abilities. These two issues limit the continued adoption of open-source models.

    To alleviate the cost of fine-tuning, many have focused on improving the hardware \cite{jouppi2023tpu} or the algorithm of the training process \cite{hu2021lora}. To improve evaluation, many have introduced more sophisticated benchmarks \cite{liang2023holistic} or rely on automated evaluation using more powerful LLMs \cite{alpaca_eval}. However, one prominent limitation faced by these approaches is that they do not analyze how the training data can be manipulated to induce a better-performing trained model.
    
    By focusing on the problem of coreset selection, we instead take a data-centric approach using influence functions. Influence functions are a method from robust statistics \cite{hampel1974infl} and are first adapted to neural-network-based machine learning models by \cite{koh2020understanding}. They approximate how much the model's prediction changes when a particular training point is removed from the training process via the model gradients. 

     
    In this work, we aim to show that Influence Functions are versatile statistical tools that can address two questions on data suitability: \textbf{1) What training points are suitable for the test set? and 2) What test points are suitable given that the model was already trained on a particular training set?} We use influence functions to identify influential data to address the two aforementioned issues by making fine-tuning more efficient by using less data and identifying test points well-covered by the fine-tuned model. 
    
    Our method is orthogonal to concurrent efforts to make fine-tuning more efficient for LLMs that focus on modifying the model architecture, such as Simplifying Transformer Blocks \cite{he2023simplifying}, or that focus on reducing the number of trainable parameters, such as LoRA \cite{hu2021lora}. It can therefore be combined with these methods. In particular, our method relies on DataInf~\cite{kwon2023datainf}, which works on top of LoRA to calculate influence values.
    
    Our main contributions can be summarized as follows:
    
    \begin{itemize}
        \item We developed \texttt{In2Core}, an algorithm that significantly reduces the size of the training set while creating a model that outperforms the one trained on the full training set.
        
        \item With \texttt{In2Core}, we can identify whether a testing point is well covered by the training set, thus providing an interpretable explanation of how a given model reacts to a particular test point.
        
        \item We further improve the efficiency of the influence function algorithm by limiting the number of model layers in calculating influence values, and we introduce a method to select the optimal number of layers given a memory budget.
        
    \end{itemize}
    
\section{Related Work}

    \subsection{Influence Functions}
        Influence functions, a subset of Data Attribution methods, seek to measure the effect of a given training point/s on a trained model. For a given training point, they seek to capture the change in behavior of a trained statistical model had that single training point not been part of the training dataset (leave-one-out-retraining). It outputs a value, called the influence value, for each training point in question.  For further discussion on Data Attribution and Influence Functions in machine learning, we refer to \citet{Hammoudeh:2022:InfluenceSurvey}.
        
        The current trend of scaling LLMs to an order of billions of parameters, as well as leveraging huge amounts of training data, pose additional computational challenges for existing influence function methods \cite{grosse2023studying, koh2020understanding}. Recent works seek to adapt data attribution for these kinds of models: TRAK \cite{park2023trak} uses an "influence function-style" estimation that simplifies the Hessian matrix, while DataInf \cite{kwon2023datainf} leverages LoRA \cite{hu2021lora} to calculate influence values for fine-tuned LLMs. These methods make data attribution tractable for modern LLMs. In our study, we use DataInf because our focus is on the fine-tuning phase of training.
        
    \subsection{Coreset Selection for LLMs}
    
        Coreset selection aims to select a subset of the full training data, such that a model well-fitted to the coreset also fits to the full data. One of the first to study it for big data viewed it as a compression method for large datasets~\cite{phillips2016coresets}. In machine learning, \citet{mirzasoleiman2020coresets} introduced a coreset algorithm to boost iterative gradient-based training. Coreset selection has been recently explored for LLMs, given the obvious problem of expensive training \cite{li2024shot, chen2024alpagasus} or to understand model properties such as in-context learning \cite{han2023understanding}. Another line of work such as \citet{han2023understanding} and \citet{wang2023farewell} apply coreset selection for selecting the pretraining data of language models. Our work differs by focusing on the fine-tuning data, which is relevant to practitioners and to the open-source community.
    
    \subsection{Coreset Selection with Influence Functions}
        Our work is closest to \citet{wang2023farewell, yang2023dataset, xia2024less} in that they use influence functions to rank data in their selection algorithms. \citet{wang2023farewell} on the pretraining data, while ours focus on the fine-tuning data. \citet{yang2023dataset} apply it for vision tasks with clearly-defined evaluation metrics (e.g. image classification) for which they can specify an objective function to achieve that metric, whereas instruction-following does not have a clear evaluation metric and remains an open-problem in the field. We use the perplexity as a coarse proxy to capture what it means to follow instructions effectively. \citet{xia2024less} also uses a definition of influence to select coresets for instruction-tuning. However, we further show that influence functions can be used to measure how much a trained model's coverage on unseen test samples. Additionally, earlier work \cite{guo2021fastif} used influence values on new, but similar training data to further fine-tune a tuned model.

\section{Influence Functions for Coreset Selection}
    In this section, we explain what influence functions are, their use in assigning a value for each training datapoint in In2Core, and an algorithmic improvement to use less memory for the influence function algorithm we use, namely DataInf \cite{kwon2023datainf}. We then explain the In2Core algorithm and perform experiments to showcase the performance of different subset selections of the data. Finally, we discuss the implications of having a smaller and better-quality training dataset, namely faster training and comparable performance.

    \subsection{Preliminaries}
    Following the formulations of \cite{koh2020understanding, Cook1980CharacterizationsOA}, an influence function measures the impact of a given training point on the change in model parameters for a given validation point. It measures this impact by up-weighing this training point and measuring the rate of change in the model parameters. 
    
    In the context of machine learning, given a model $f$ parameterized by the empirical risk minimizer $\theta*$, we consider how a given training point changes the validation loss. Given a training point $z$ and a validation point $z'$, the influence of $z$ on $f(\theta*)$'s performance on $z'$ is defined as
    \begin{equation}\label{e:inf}
        I(z, z') := (\nabla_{\theta} \loss(z') |_{\theta=\theta*})^T I_{\theta*} (z)
    \end{equation}
    
    where \begin{equation}\label{e:hessian}
        I_{\theta*} (z) := \frac{d\theta^z}{d\epsilon} |_{\epsilon=0} = -H(\theta*)^{-1} \nabla_{\theta} \loss_k
    \end{equation} 
    
    and $H:= \nabla^2_\theta (n^{-1} \sum^n_{i=1} \loss(f_{\theta} (z) )$, the Hessian of the empirical loss.
    
    Equation ~\eqref{e:inf} can be extended to an entire validation set $D^{val} := {z'_{i}}^m_{i=1}$ by taking the average gradient of the validation set as
    
    \begin{equation}\label{e:inf_all}
        I(z, D(^{val}) := (\frac{1}{m} \sum^m_{i=1} \nabla_{\theta} \loss(z'_{i}) |_{\theta=\theta*})^T I_{\theta*} (z)
    \end{equation}
    
    The influence value and its sign represents the impact of including a particular point in the training set on a given validation point/set, via the validation loss. For clarity, we follow \cite{pruthi2020estimating} in using their terminologies. We define
    \begin{enumerate}
        \item \textbf{Proponents} - Points with negative influence values. Their addition to the training set reduces the validation loss.
        \item  \textbf{Opponents} - Points with positive influence values. Their addition to the training set increases the validation loss.
    \end{enumerate}
    
    For current language models, which rely on transformer-based models with billions of parameters, computing the Hessian term in Equation \ref{e:hessian} is extremely expensive. In practice, we rely on the Hessian's estimation. In this paper, we use the estimation of influence from DataInf by \cite{kwon2023datainf}.

\begin{figure}[t]
        \centering
        \includegraphics[width=0.48\textwidth]{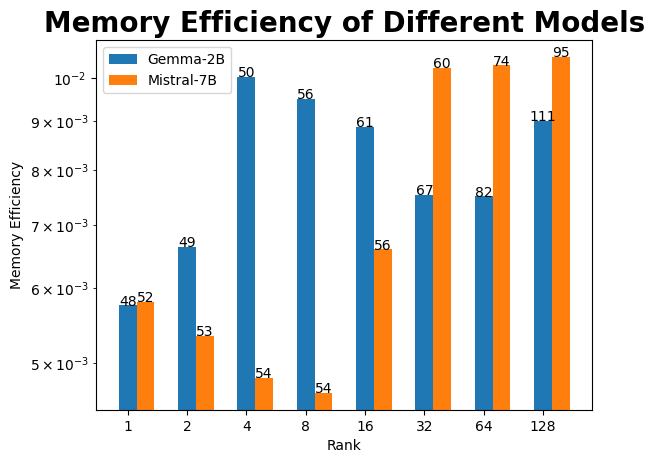}
        \caption{Memory efficiency across various number of LoRA layers to consider when calculating influence values. The numbers above each bar correspond to the virtual CPU memory consumed (in GB). The memory efficiency rate is not linear, and varies across different models. One should sample a subset to calculate the optimal combination of CPU virtual memory and number of layers to use given their hardware constraints.}
        \label{fig:mem_eff}
    \end{figure}

    \noindent\textbf{Efficient DataInf}.  
    \label{sec:improv}
    While the DataInf algorithm efficiently calculates the gradients of the model parameters for each point, in practice the CPU memory consumption from the gradient collection becomes a bottleneck as the size of the training dataset grows. This is a problem for sizeable fine-tuning datasets such as the Ultrachat subset we use, which contains roughly 50,000 training points. Figure \ref{fig:mem_eff} shows the trend for N=250 points. 
    
    We found that selecting the number of layers is non-trivial. To illustrate this, let \textit{all-layer} denote the case where we use every layer with a LoRA adapter. First, using large $k$ can result in marginal gains at approximating influence values from all-layer at the expense of the memory budget. Second, different model architectures give different efficiency, implying $k$ is different for each model (and dataset). These two reasons require a hyperparameter search for $k$.
    
    To enable comparisons across different models given the same dataset, we define the metric Memory Efficiency, $s := \frac{\rho}{\text{number of layers used}}$, where $\rho$ is Spearman's rank correlation coefficient of a particular setup with respect to the case where all layers are used to calculate the gradient (\textit{all-layer}). $\rho$ is a metric of interest because it describes how faithful the ranking of the current setup is to the ranking in the all-layer setup. Note that $k$ layers refer to the first $k$ layers because the first layers capture influence better than the last layers \cite{yeh2022better}. We then simply set $k$ as the number of layers with the highest $s$, subject to our virtual memory budget. For our experiments, we used $k=\{8, 16\}$ for Gemma-2B and Mistral-7B, respectively. Practitioners should perform this preliminary evaluation of memory efficiency on a small subset, in order to avoid out-of-memory (OOM) errors when calculating the influence values.

\subsection{Algorithm}
    
    \begin{figure*}
        \centering
        \includegraphics[width=1\textwidth]{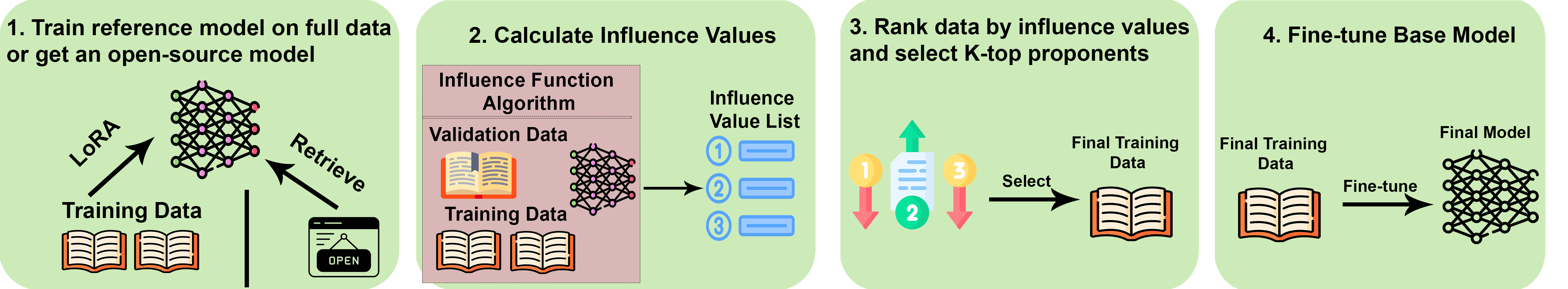}
        \caption{Overview of In2Core for coreset selection. From left to right, we first calculate the influence values of each training point using the validation dataset and a reference model fine-tuned on the full dataset with LoRA. Then, we rank the training points by influence values and select the $h$ highest-scoring proponents as the final training data, where $h$ is a hyperparameter. Finally, we train a base model on this final training data. Both the reference and base model may have distinct architectures from each other.}
        \label{fig:coreset_algo}
    \end{figure*}
    
Given a large dataset \full, evaluation set \eval, and reference model $f$ fine-tuned on \full, to select a coreset $D_p$, where $D_p \subset$ \full
\begin{enumerate}
    \item Calculate the influence of each $z\in$ \full using $f$.
    \item Rank each $z$ by influence value.
    \item Select $p$ and get the top-$p$ proponents as elements of $D_p$.
\end{enumerate}

The first step is the most expensive step as the influence function algorithm visits each point. In particular, it is necessary to get the model gradients for each point, which creates a memory bottleneck. In Section \ref{sec:improv}, we improved the influence function algorithm we use to mitigate this.

The second step provides an ordered list of the training data with respect to each points' influence value. For the third step, note that $p$ is a hyperparameter and depends on the training budget of how many data points the training can accommodate. Given the definition of proponents, the "top" proponents are the points with most negative influence values. 

\subsection{Experimental Setup}
For coreset selection, we use Mistral-7B-v0.1 \cite{jiang2023mistral} \& Gemma-2B \cite{geminiteam2023gemini} as base models and fine-tune them on subsets of a "full" training dataset, which is a random 50k subset of the first round of dialogues from Ultrachat-200k \cite{ding2023enhancing}. Note that the models we use are allowed for research. We evaluate the fine-tuned models on a disjoint 250-random subset from Ultrachat-200k, following the same format as the training dataset. For both Sections \ref{eval_coreset} and \ref{eval_cov}, we use perplexity \cite{jelinek1977perplexity} on the evaluation set to measure model performance, given that it is directly related to the loss, whose reduction is the goal of fine-tuning (training loss) and is a component to the calculation of influence values (validation loss).

For simplicity, we refer to the mean perplexity of the model on the validation dataset as perplexity and BERTScore between each validation point and the training dataset as similarity.

\begin{figure*}[t]
     \centering
     \begin{subfigure}{0.3\textwidth}
         \centering
    \includegraphics[width=1.0\textwidth]{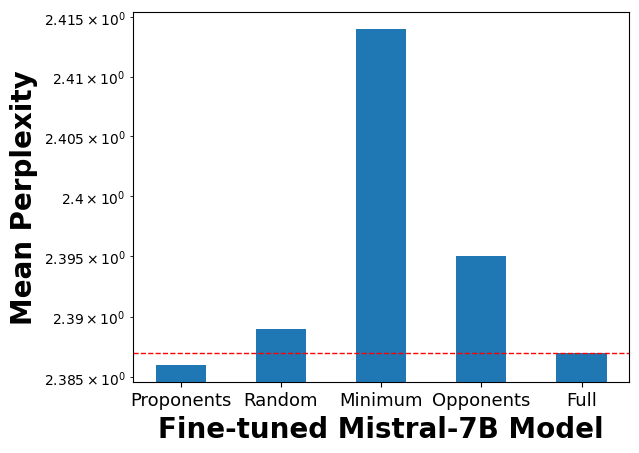}
    \caption{Performance of Models (Ref: Mistral-7B)}
     \end{subfigure}
     \hfill
     \begin{subfigure}{0.3\textwidth}
         \centering
         \includegraphics[width=1.0\textwidth]{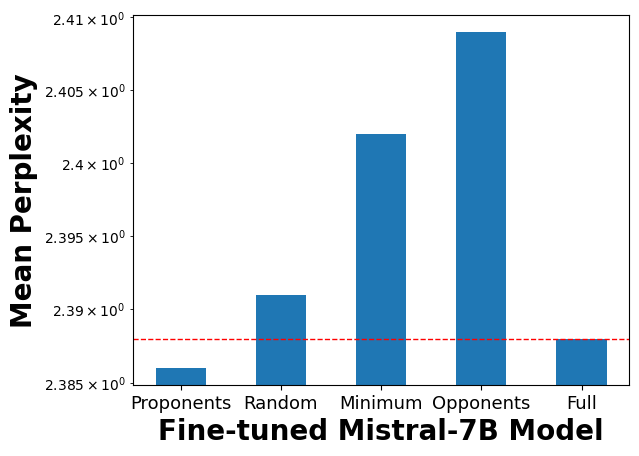}
    \caption{Performance of Models (Ref: Gemma-2B}
     \end{subfigure}
     \hfill
     \begin{subfigure}{0.3\textwidth}
         \centering
    \includegraphics[width=1.0\textwidth]{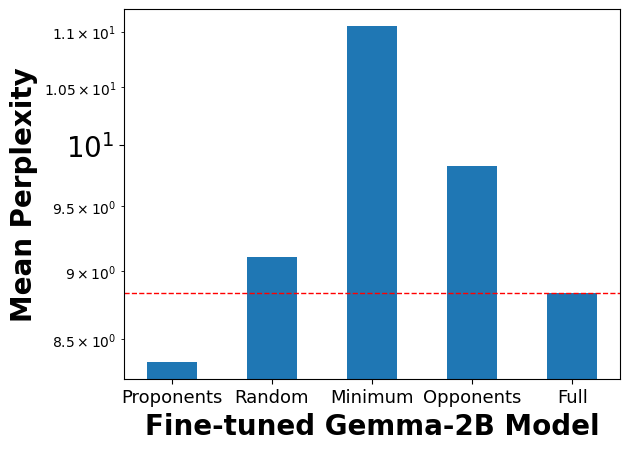}
        \caption{Performance of Models (Ref: Gemma-2B}
     \end{subfigure}
        \caption{Mean Perplexity of models fine-tuned on different coreset selection strategies on the 250-point Ultrachat Evaluation Set. These strategies differ by selecting based on the influence values. 'Full' denotes a model trained on the full training data. Proponents, which is the default strategy in practice, outperform all groups except Full in all cases. Interestingly, some strategies result in a worse model than Random. In Section \ref{disc_coreset}, we argue that some points inherently degrade model training when included (e.g. Minimum and Opponents).}
        \label{fig:coreset}
\end{figure*}

\subsection{Results}
\label{eval_coreset}
\noindent\textbf{Model Perplexity}.
Using Equation ~\eqref{e:inf_all}, we calculate the influence values with respect to the entire evaluation dataset. Figure \ref{fig:coreset} illustrates our algorithm's performance at coreset selection. The x-axis represents the different selection strategies based on their influence values. Minimum refers to selecting points whose influence values are nearest to zero in absolute terms. Random refers to uniformly-sampled points. In all figures, selecting Proponents consistently leads to the lowest perplexity among the other subset selection strategies. This applies when, with respect to the base models being fine-tuned, the reference model is smaller (as in the case with Gemma-2B or it is of a different architecture with a different pretraining data (as with Phi-2 pretrained on synthetic Textbook-quality data \cite{li2023textbooks}). Importantly, selecting the best $h=25$k proponents, which is half of the full dataset, leads to a model that has lower perplexity than a model trained on the full dataset. This shows that it is possible to achieve better model performance with less training data.

\begin{figure}[t]
    \centering
    \includegraphics[width=0.48\textwidth]{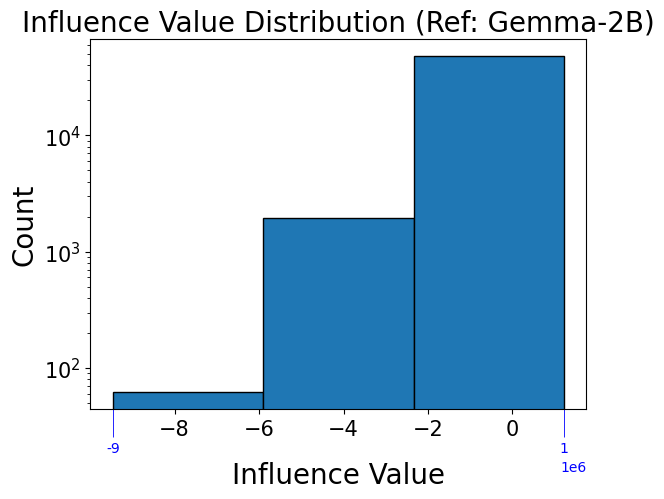}
    \caption{Histogram of influence values of the 50k training set using Gemma-2B as the reference model. The distribution is left-skewed, with the majority of influence values being negative. The Opponent and Minimum groups, (points with positive values and values smallest in influence value magnitude respectively) are located to the right side of the histogram and have high overlap (overlap coefficient = 0.59. Note that the numbers in blue denote the rounded value of the extremes of the distribution.}
    \label{fig:inf_vals_hist}
\end{figure}

Interestingly, the behavior of Minimum and Opponents are similar. In some cases selecting Minimum leads to a model with higher perplexity than Opponents. The similarity in behavior is explained by the distribution of influence values, which is left-skewed as shown in Figure \ref{fig:inf_vals_hist}. Thus, there is a significant overlap between the points selected for Minimum and those for Opponents. For how Minimum can lead to higher perplexity than Opponents, it may indicate that points with the least absolute influence changes the model's behavior the least compared to other strategies. In other words, it best retains the model's behavior before fine-tuning, which in our experiments is a base model that is trained to produce verbose output. A high perplexity here is then explained by the Minimum model producing longer text than the other models.

\begin{table}[t]
\centering
\begin{tabular}{|l|l|}
\hline
\textbf{Model} & \textbf{Accuracy} \\ \hline
Full & 0.32 \\ \hline
Proponents-25k & 0.30 \\ \hline
\end{tabular}
\caption{MMLU Average Accuracy (Zero-shot learning). All numbers are rounded-off to 2 decimal places.}
\label{table:mmlu}
\end{table}

\noindent\textbf{MMLU Accuracy}. 
We further evaluate the two models of interest: Full (trained on the full dataset) and Proponents-25k (trained on the best 25k proponents) on the Massive Multitask Language Understanding (MMLU). MMLU tests language models for extensive world knowledge and problem solving capabilities in multiple-choice format ~\cite{hendrycks2021measuring}. We use the Gemma-2B models fine-tuned on the Ultrachat subset, which are the same models in the previous experiments. Table \ref{table:mmlu} reports the average accuracy across the different subjects as the fine-tuning was intended to improve general instruction-following.

Proponents-25k surprisingly performs similarly to Full for a wide range of tasks, and attains a similar accuracy as Full. There may be large differences in some subjects (around $10-25\%$), but there is similar performance in capabilities which those subjects are components of. For example, for \textit{logical reasoning}, which is composed of subjects such as \{\textit{formal logic, elementary mathematics, logical fallacies}\}, Full performs better in the first two subjects, but Proponents-25k performs better for the last subject. We postulate this is because our evaluation set represents samples for general capabilities. We expect In2Core can be used to improve specific capabilities if the evaluation set's distribution matches those capabilities. While it performs slightly worse, Proponents-25k being trained on half the data provides further evidence that additional data used by Full is only marginally useful. 

\subsection{Discussion}
\label{disc_coreset}
\emph{More data can result in worse models.} For a given test set, some points are \textbf{inherently harmful} to include in the training data. This goes against the conventional idea that more data is better for transformer-based LLMs, given how their typical training (both pre-training and fine-tuning) is inherently data-intensive. Our work sheds light on the fact that not all points are of equal quality, and we can dramatically reduce the size of the training dataset if we only keep data that works towards our metric of interest. How specifically these Opponents, when added to the training dataset, can undo the influence of Proponents is not well-understood, but this may be related to the phenomenon of catastrophic forgetting \cite{goodfellow2015empirical}, where the abilities learned by a model from a given proponent may be forgotten when learning from an opponent/s.

We expect such harmful points to become increasingly common as synthetic data is increasingly used to fine-tune LLMs, where there is risk of model collapse from using poor-quality training data \cite{shumailov2023curse}, or a training dataset that does not accurately reflect the distribution of the validation dataset. As such coreset selection will become an increasingly important method for most types of LLM training. Our algorithm can be incorporated as a pre-processing step for practitioners before fine-tuning.

\emph{Influence values from smaller models transfer to larger models.} Surprisingly, we find that a smaller model (in this case almost 4x smaller than the model to be fine-tuned) can act as a reliable reference model, specifically for selecting what points to avoid adding in the training dataset. We expect that this transferring ability will remain as long as the reference model sufficiently learns the underlying training distribution after training. Importantly, this allows a further cost reduction of the two most expensive steps in our algorithm -- namely the fine-tuning of the reference model on the full training dataset and the subsequent gradient collection.

\emph{Influence values transfer across model architectures.} We show that our method applies to reference models with distinct model families, even those with completely different pre-training regimes. This implies that a practitioner can simply bypass training a reference model if an open-source version is available.

\section{In2Core for Model Coverage}
In this section, we use In2Core at test-time. Specifically, we use influence values to measure how "suitable" individual test points are to a final model trained on a given training set. We first describe the rationale and our method for analyzing model coverage. Then, we perform an experiment comparing our method to analyzing model coverage via semantic similarity. Finally, we discuss the implications of the experiment. Note that we use the Gemma-2B model fine-tuned on the Proponents dataset for our experiments, described in \ref{eval_coreset}, and thus omit an Experimental Setup section.

Evaluating a trained instruction-tuned model is difficult considering the space of acceptable outputs given certain questions. Rather than solely looking at discrete outputs to determine if the model generalizes to a test point, we instead augment this analysis by also looking at the training set. We capture the influence of the entire training set by taking its average gradient, thereby giving a single influence value for each test point. This is akin to measuring how much a test point falls within the training distribution. To the best of our knowledge, this is the first time that influence values have been used in the literature to analyze how a given training set as a whole is appropriate for a given test point via a trained model. Importantly, this method is compatible with different evaluation metrics. In Section \ref{eval_cov}, we specifically apply this on the perplexity metric.

For instruction-tuned LLMs, having an indication on whether a given test input fits within its training distribution is essential because there is either an absence of a ground-truth to validate with or a ground-truth that is hard to articulate (e.g. a writing task where there is a space of acceptable outputs). We would like to emphasize that influence functions should not be used by themselves to evaluate a model's capability. It may be the case that the entire training set is beneficial to the model at learning a specific capability, but the training set size is insufficient for the model to learn that capability. Rather, influence functions provide further guidance at understanding model capability at evaluation by tying that capability to the training set, especially when one wishes to determine how to improve the model (as in coreset selection). Our method can be integrated in pipelines for model debugging (\textit{which test points does the model need more data for?}) and handling inference requests (\textit{which requests is the model not suited to perform well given its training?}) in production-grade LLMs.

Furthermore, using influence functions in this manner can be computationally-efficient. We can exploit a feature of influence functions to our advantage: the most expensive computational step is calculating the gradients, but once the gradient computation is done, we can store them. In particular, we store the average gradient of the training dataset, and we can arbitrarily use it for future inference on unseen test points. While we still need to compute the model gradients with respect to new test points, the largest cost, coming from the training set, is only performed once. 

\subsection{Results}
\label{eval_cov}
\begin{figure}[t]
     \centering
     \begin{subfigure}{0.45\textwidth}
         \centering
         \includegraphics[width=1.0\textwidth]{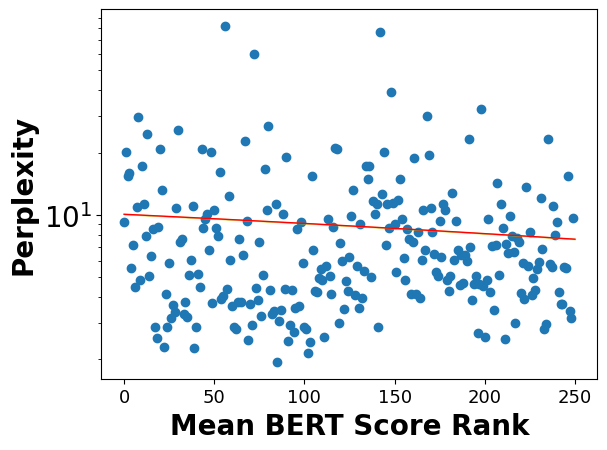}
         \caption{Similarity of the Training Set on Evaluation Set}
     \end{subfigure}
     \begin{subfigure}{0.45\textwidth}
         \centering
         \includegraphics[width=1.0\textwidth]{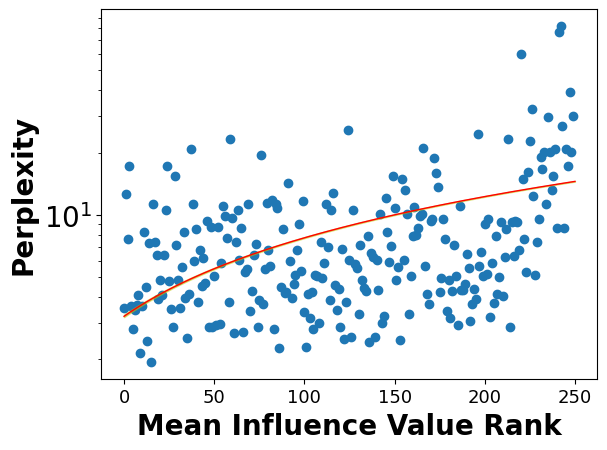}
         \caption{Influence of the Training Set on Evaluation Set}
     \end{subfigure}
        \caption{Relationship between Perplexity and Measures of Importance of the Training Set on the Test Set (N=250). Influence values provide a better signal (correlation coefficient = 0.56) to indicate how well a model generalizes to a particular test point compared to semantic similarity (correlation coefficient = -0.087).}
        \label{fig:gen_ppl}
\end{figure}

In Figure \ref{fig:gen_ppl}, we compare two different notions of measuring the importance of the entire training set to each evaluation point: Semantic similarity via BERTScore \cite{zhang2020bertscore} and Influence via influence functions from DataInf. The BERTScore of two points is their cosine similarity in the contextual embedding space of BERT \cite{devlin2019bert}. BERTScore aims to capture the semantic similarity between two different sequences of texts. In our experiments, we use DistilBERT \cite{sanh2020distilbert} for efficiency. For this case, we calculate the BERTScore of each training point to each test point. Then for each test point, we take the mean of the scores of all training points on that test point. We define this as the similarity score of the entire training set for that particular test point. Therefore for each test point, we obtain a scalar value as the similarity score. Meanwhile, to calculate the influence of the entire training set on each test point, we take the mean of the training point gradients to reduce it into a single point. We then calculate influence values with respect to individual test points as per normal.

For both types of importance, we sort each test point into ascending order and plot the rank of each point. We compare these ranks to the perplexity of the model on the test point, where the model is the Gemma-2B fine-tuned on the entire training set. The red line for each graph denotes the linear regression line. Coefficient values are within a 95\% confidence interval, but the intervals are too narrow to visualize. For semantic similarity as a metric, we hypothesize perplexity and semantic similarity to have a negative linear correlation (i.e. a downward trend) because test points with lower semantic similarity to the training set signifies that the test points are from a different distribution than the training set. However, while we observe this downward trend, it is a very weak correlation (coefficient = -0.087), suggesting that semantic similarity as measured by BERTScore is not a strong signal to indicate the fit of the training set to the test points.

In contrast, perplexity and influence values exhibit a stronger correlation. For Influence as a metric, we hypothesise perplexity and Influence to have a positive linear correlation (i.e. an upward trend) because test points assigned negative influence values imply that the training set is a proponent of those test points, and the training set is an opponent for the reverse case. We see this relationship between perplexity and influence, and observe that the correlation is stronger (coefficient = 0.56). This suggests that using influence values provide a stronger signal to indicate whether the model can generalize to a particular test point. Furthermore, using influence values is cheaper compared to using BERTScore, because we can reduce the training set to a single point when representing it as a gradient, compared to calculating point-wise similarities with BERTScore.

\subsection{Discussion}
\label{disc_cov}
\emph{Influence functions and evaluation metrics are complementary and co-dependent at improving model capability.} 
We claim that these two measures are complementary because they provide an actionable feedback loop at improving the model. Evaluation metrics measure how close the model is behaving towards a desired behavior, and influence functions verifies if the training data does improve the model towards that desired behavior. Our coreset selection aglorithm is an example of how the two measures synergize instruction-tuning, which in our experiments focus on perplexity and influence. We claim that they are co-dependent because using these measurements on their own provide practitioners inadequate information on (a) whether the training data is sufficient at improving a capability, or (b) whether the additional training data will even contribute towards improving the capability, respectively.

\emph{Measuring semantic similarity of training points is insufficient to capture a model's ability to generalize to a test point.}
A model's ability to generalize to a test point is dependent on the model's loss on that test point. Semantic similarity, while providing an intuitive explanation for model generalization (a model learning from similar points will perform well on an unseen but similar point), does not completely capture different ways how a model learns from different training points to generalize to a test point. In particular, some points may not be semantically related to the test point, but they still serve to reduce the loss on that point. Since influence values are calculated with respect to the loss value (e.g. how much a given training point reduces the loss on an evaluation point/set), they are a closer metric to capture model generalization, and Section \ref{eval_cov} demonstrates this empirically via the correlation coefficients.

\section{Conclusion}
In this work, we have shown that blindly adding data into the training set can \textit{degrade} the model's performance. We developed In2Core, an algorithm for coreset selection based on influence functions. Our algorithm is practical and can be combined with other pre-processing steps to make supervised fine-tuning more efficient. We show that training on just half of the original training data can be comparable with the performance of a model trained on the full data for a variety of reference and target models. Importantly, (1) we observe a transfer effect where a reference model of a different architecture can be used to select the coreset data, and (2) this effect holds even with a reference model that is smaller than the model we are fine-tuning. Finally, we explore the use of influence to measure a trained model's coverage. We show that measuring data influence and an evaluation metric can aid in efficient model training than merely using an evaluation metric alone. We also show that this "influence" computed via influence functions captures a model's coverage better than by semantic similarity.


\section*{Limitations}

\textbf{LLM Evaluation}
There is ongoing work to evaluate instruction-following models on a variety of tasks \cite{hendrycks2021measuring,eval-harness,liang2023holistic,open-llm-leaderboard,wang-etal-2024-seaeval,wang2024resilience}, but reaching a consensus on how to best evaluate such general-purpose models is still an open issue. Furthermore, there is an existing line of work that uses a larger model for automatic evaluation \cite{alpaca_eval,vicuna2023,zhao2024preliminary,chen2024alpagasus,wang2024audiobench}, most popularly GPT-4 \cite{openai2024gpt4}. However, the decision of the larger model can be biased to models from the same family. \citet{alpaca_eval} Furthermore, the output is opaque and uninterpretable because there is no guarantee that the accompanying explanation to its answer is faithful~\cite{jacovi2020towards}.

\noindent\textbf{Group influence}.
Our formulation of influence looks at the contribution of each points individually, but not their contribution as a group of points. We assumed that removing / adding groups of points based on their \textit{individual} influences would create the expected effect (e.g. a sufficient learned model). Although we demonstrated this empirically in Section \ref{eval_coreset}, we leave to future work the theoretical analysis if there are better methods to select points based on their individual or group influences.

As for selecting a group of points based on their group influences, we would like to stress that our algorithm could be applied for group influence by replacing DataInf with the group influence algorithm of choice. Calculating group influence with influence functions has first been studied by \cite{koh2019accuracy} and algorithms to calculate it are becoming more efficient~\cite{basu2020secondorder}.

\noindent\textbf{Average gradient of the tokens as the gradient of the entire sequence}.
We take the average gradient of the tokens when calculating the gradient of each point to make influence value calculations tractable. \citet{xia2024less} showed that this method penalizes the influence of points with long sequences of text. This is a fundamental limitation for definitions of influence that involve calculating the gradient of sequences of varying lengths. See Table \ref{fig:seq_len} for training examples selected by In2Core. 

\section{Ethics Statement}
Training LLMs is compute-intensive and thus carbon emissions-heavy~\cite{rillig2023risks}. Our paper introduces a method to make LLM fine-tuning more efficient, which can lead to positive environmental impact. Using less data to fine-tune translates to less electricity and water consumption, all else being equal. We intend our method to be adopted by practitioners to reduce the cost of training. Our results are also transparent and derived from open-source LLMs \& datasets.

\section*{Acknowledgements}
This research is supported by the National Research Foundation, Prime Minister’s Office, Singapore under its Campus for Research Excellence and Technological Enterprise (CREATE) programme. The computational work for this article was partially performed on resources of the National Supercomputing Centre, Singapore (https://www.nscc.sg).

\bibliography{anthology,custom}
\bibliographystyle{acl_natbib}

\appendix
\section{Appendix}
\label{sec:appendix}
\subsection{Experimental Hardware Details}
We used NVIDIA A100 Tensor Core GPUs. For influence values computation with the reference models, we used 1 GPU. For fine-tuning of models trained on the full dataset, we use 4 GPUs. For the rest of the fine-tuning setups, we used at most 2 GPUs. We estimate that we used 100 - 200 GPU hours on this project.

\subsection{Coreset selection for Llama2-7B}
We conduct further experiments by using Llama2-7B \cite{touvron2023llama} as the target models to be fine-tuned. While the proponents version where Gemma-2B is the reference model performs better than the full version, the proponents version where a Llama2-7B is the reference model performs worse than the full version. The latter is likely limited by the number of layers used to calculate the influence values, because the highest spearman correlation we can attain given our hardware constraints was less than 50\% for Llama2-7B.

\begin{figure}[hbt]
     \centering
     \begin{subfigure}{0.45\textwidth}
         \centering
         \includegraphics[width=1.0\textwidth]{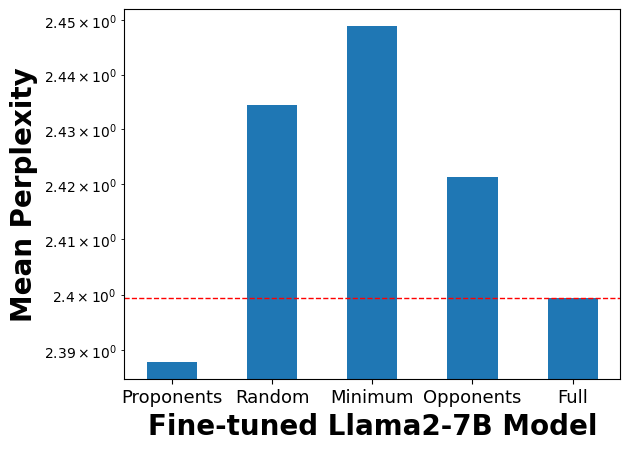}
         \caption{Performance of Models (Ref: Gemma-2B)}
     \end{subfigure}
     \begin{subfigure}{0.45\textwidth}
         \centering
         \includegraphics[width=1.0\textwidth]{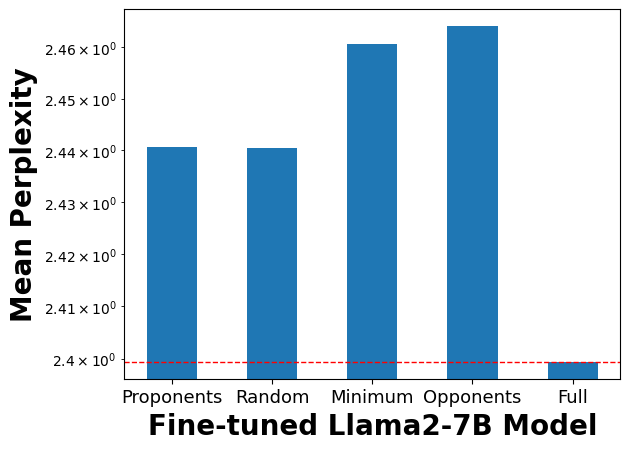}
         \caption{Performance of Models (Ref: Mistral-7B)}
     \end{subfigure}
        \caption{Mean Perplexity of models fine-tuned on different coreset selection strategies on the 250-point Ultrachat Evaluation Set for Llama2-7B models. 'Full' denotes a model trained on the full training data.}
\end{figure}

\begin{figure}[hbt]
    \centering
    \includegraphics[width=0.48\textwidth]{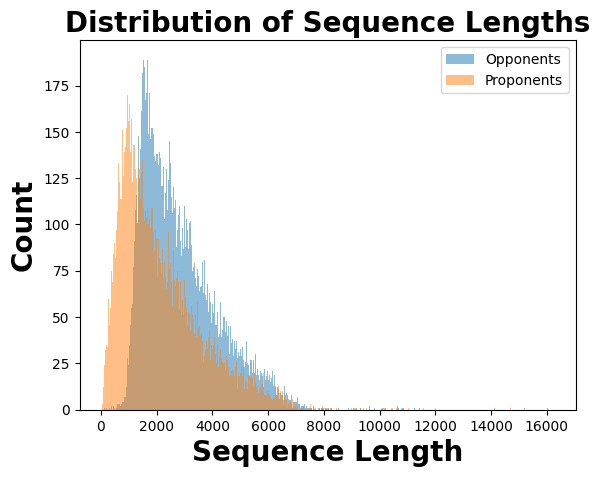}
    \caption{Influece functions are biased towards assigning higher scores to longer sequences. This stems from the fact that the value calculations are based on taking the average gradient of the token as the sequence gradient.}
    \label{fig:seq_len}
\end{figure}

\end{document}